\def\year{2019}\relax

\documentclass[letterpaper]{article} 
\usepackage{aaai19}  
\usepackage{times}  
\usepackage{helvet} 
\usepackage{courier}  
\usepackage[hyphens]{url}  
\usepackage{graphicx} 
\usepackage{graphicx}  
\usepackage{todonotes}
\usepackage[caption=false]{subfig}

\title{Enabling Intuitive Human-Robot Teaming Using Augmented Reality and Gesture Control}

\author{\Large \textbf{Jason M. Gregory\textsuperscript{\rm 1}, Christopher Reardon\textsuperscript{\rm 1}, Kevin Lee\textsuperscript{\rm 2}, Geoffrey White\textsuperscript{\rm 3}, Ki Ng\textsuperscript{\rm 3}, Caitlyn Sims\textsuperscript{\rm 4}} \\ 
\textsuperscript{\rm 1} US CCDC Army Research Laboratory, Adelphi, MD, USA \\
\{jason.m.gregory1, christopher.m.reardon3\}.civ@mail.mil \\
\textsuperscript{\rm 2} US CCDC ARL Research Associateship Program administered by Oak Ridge Associated Universities \\
klee23456@gmail.com \\
\textsuperscript{\rm 3} Defence Science and Technology, Melbourne, Australia \\
\{geoffrey.white, ki.ng\}@dst.defence.gov.au \\
\textsuperscript{\rm 4} DST IEP Program with Swinburne University of Technology \\
caitlyn.sims@dst.defence.gov.au \\
}

\begin{document}

\maketitle

\begin{abstract}
Human-robot teaming offers great potential because of the opportunities to combine strengths of heterogeneous agents. However, one of the critical challenges in realizing an effective human-robot team is efficient information exchange - both from the human to the robot as well as from the robot to the human. In this work, we present and analyze an augmented reality-enabled, gesture-based system that supports intuitive human-robot teaming through improved information exchange. Our proposed system requires no external instrumentation aside from human-wearable devices and shows promise of real-world applicability for service-oriented missions. Additionally, we present preliminary results from a pilot study with human participants, and highlight lessons learned and open research questions that may help direct future development, fielding, and experimentation of autonomous HRI systems.
\end{abstract}

\section{Introduction}
\label{sec:intro}
Human-robot teaming has become the focus of much research because it offers great potential for completing various service-oriented missions more robustly and efficiently by combining the strengths of heterogeneous agents. Inherently, human-robot teaming encompasses both human-robot interaction (HRI) and shared mission execution as it requires agents to coordinate, cooperate, and collaborate in order to achieve a common goal. One of the core challenges to realizing effective teaming is improved information exchange between people and robots because of various mission requirements including correct mental models, calibrated trust, and enhanced situational awareness \cite{Szafir2019}. The challenge of team-focused communication is especially complex because it requires consideration of information flow in both directions, i.e., from the human to the robot as well as from the robot to the human. 

Information flow from the human to the robot is critical to teaming because it provides the human with a way to inform the robot of the prioritized mission goals that may dynamically change over time as new information is acquired. This often times is in the form of commands and, as a result, necessitates solutions that enable fast, expressive, natural communications that convey meaningful task information. Previous efforts have considered several intuitive modalities for human-to-robot communication; including, natural language \cite{Huang2019}, gestures \cite{Elliott2016}, and multi-modal communication \cite{Cacace2016}. Ultimately, streamlined interactions allow for the human to have greater control over their robotic teammates and ideally should be as effortless as communicating with another human counterpart. 

In the other direction, information flow from the robot to the human provides enhanced situational awareness so that high-level decisions made by the human are more informed. When equipped with sufficient artificial intelligence (AI) and resourceful decision-making capabilities, robots can autonomously collect information and perform tasks that would otherwise be too dangerous, time-consuming, or mundane for humans. The essential questions are then: what mission-critical data should be provided to the human, and how should this information be displayed? Data sharing and visualization must be well-defined and non-intrusive in order to maximize understandability and minimize the threat of overloading the human. The literature has already developed design spaces and taxonomies for working in reality, augmented reality (AR), virtual reality (VR), and mixed reality (MR) environments to optimize data visualization \cite{williams2019reality}. To this end, researchers have proposed solutions for HRI including AR in instrumented environments \cite{Walker2018}, VR for dexterous manipulation tasks \cite{Whitney2018}, and MR interfaces for deictic gesture and natural language \cite{Williams2019deictic} and human-programmed robot motion \cite{Gadre2019}.

Here, we propose a technology specifically designed to address the two-way communications challenge of human-robot teaming. We utilize a glove with gesture recognition for human-to-robot communication, and an AR head-mounted device (AR-HMD) to enable robot-to-human communication. We developed autonomous behaviors so that the robot can support collaborative missions and accept human-issued commands for dynamic, expedited operations. Since the robot projects its received commands, autonomous plans, and actions in the AR-HMD display, our system as a whole can be thought of as an MR-based solution. By building on our previous works \cite{White2018}, \cite{Reardon2018}, \cite{Reardon2019}, and marrying wearable devices with artificial intelligence and tactical data visualization, we develop a system that enables intuitive human-robot teaming with no external instrumentation and has applicability to service-oriented missions in real-world environments. Our contributions are three-fold. First, we present the hardware and software architecture of our proposed system for both the human and robotic teammates in the subsequent three sections. This includes our AI-based algorithms, proposed command set, and corresponding gestures. Second, in the Field Experiment section, we highlight preliminary results from our pilot study with human participants in order to glean insights regarding usability and show emergent use cases and modalities of our human-robot teaming system. Finally, in the Lessons Learned section, we analyze our experimental HRI field study and identify open questions for tools that support AI-focused HRI to help focus practitioners and researchers in the advancement of future technologies.



\section{System Design}
\label{sec:design} 
\begin{figure}[htbp]
  \centering
  \includegraphics[width=0.80\linewidth]{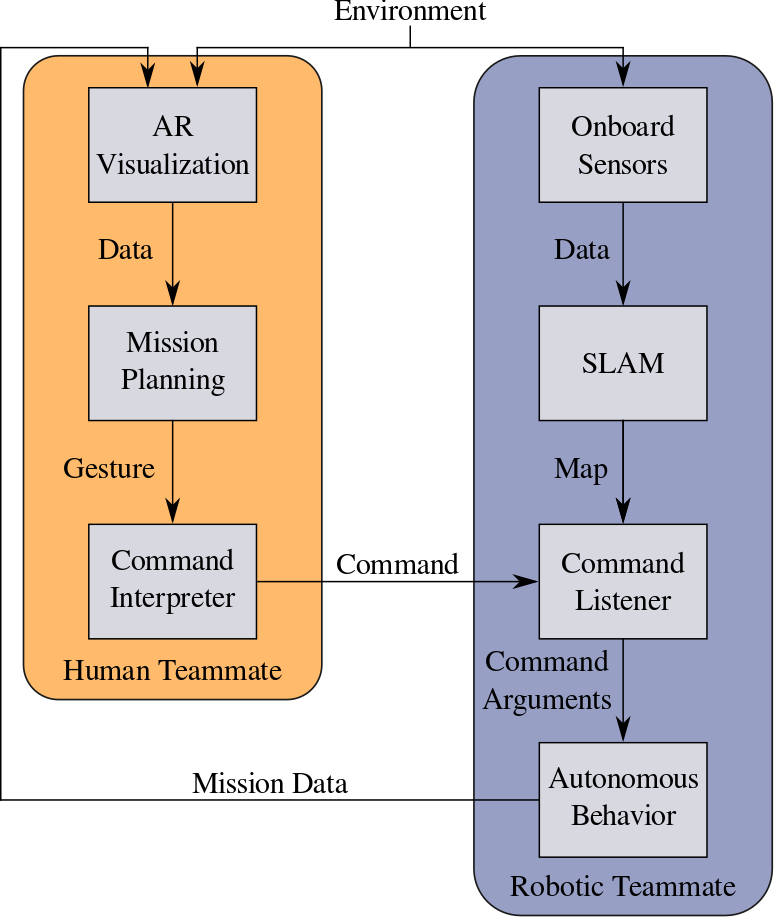} 
  \caption{System diagram of our proposed AR-enabled, gesture control system for intuitive human-robot teaming.}
  \label{fig:system} 
\end{figure}

A critical component to any fieldable, human-robot system is intuitive control. To achieve this, both the human and robot teammates must individually possess the proper hardware and autonomous decision-making capabilities that enable efficient sharing of data and two-way communication. The robot must collect and present data in a human-understandable fashion as well as provide sufficiently-autonomous behaviors in order to execute human-provided commands. This requirement necessitates the development of AI and robust autonomy. Additionally, the human must be able to issue commands in a fast and accurate manner, and have sufficient understanding of the mission requirements so that commands are relevant. By defining the necessary requirements and outfitting both the human and robot teammates, we have developed such a system that enables autonomous, cooperative exploration and simultaneous data acquisition for enhanced situational awareness with virtually no a-priori training or environmental information. It is important to note that no external instrumentation or modification of the unstructured, operational environment is required by our system. A high-level system diagram, which is explained in greater detail in the following sections, is shown in Figure~\ref{fig:system}. 

\section{Outfitting the Robotic Teammate}
\label{sec:robot}
We treat the robotic agent as a subordinate, but capable teammate for the human. In a realistic deployment, the robot will likely be tasked to achieve enhanced situational awareness, improved safety through increased standoff distances, and expedited services important to a human or larger team. Thus, it is crucial for the robot to prioritize commands issued by the human; however, it must also be able to make intelligent decisions autonomously to alleviate the human of the burden of constant supervision. In the subsequent sections, we first describe the hardware we found necessary for the robot to have access to mission-critical data, and then describe the software components we developed to enable autonomous decision-making. 

\subsection{Hardware}
\label{subsec:robot-hardware}
\begin{figure}[htbp]
  \centering
  \includegraphics[width=0.75\linewidth]{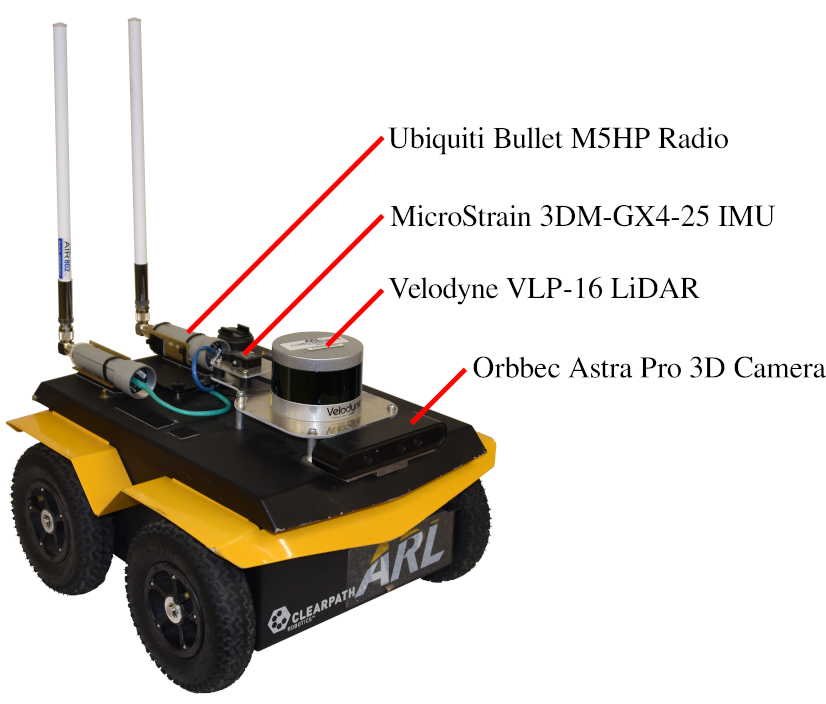}
  \caption{The configuration of the platform and sensors that serve as the robotic teammate.} 
  \label{fig:robot} 
\end{figure}

For this effort we used a Clearpath Jackal\nocite{Jackal} robot as a surrogate for a robotic teammate. This wheeled platform is $0.508 \times 0.430 \times 0.250$ m, travels at a maximum velocity of $2.0$m/s, and serves as a mobile base for our various sensors. On board the chassis is an Intel Core i5-4570TE CPU, which runs Ubuntu $14.04$ and the Robot Operating System (ROS) Jade \cite{Quigley2009ros}. We installed on the Jackal a Velodyne VLP-$16$ Light Detection and Ranging (LiDAR) sensor\nocite{Velodyne}, which has $16$ channels with a range of $100$m that collect approximately $300$,$000$ points per sec in a $360^{\circ}$ horizontal field of view and a $30^{\circ}$ vertical field of view. For improved mapping and state estimation performance, we installed a MicroStrain $3$DM-GX$4$-$25$ inertial measurement unit (IMU)\nocite{Microstrain}. To improve situational awareness and record experimental video, we installed an Orbbec Astra Pro\nocite{astra} $3$D camera that provides RGB-D data, although it was not relied upon for the autonomous behaviors in this work. For communications between the human and robot, we used a Ubiquiti Bullet M$5$HP $5$GHz WiFi radio\nocite{Bullet}. The robot, equipped with its sensors, is shown in Figure~\ref{fig:robot}.

\subsection{Autonomous Behaviors}
\label{subsec:behaviors}
The basis of our autonomous behaviors is a robust Simultaneous Localization and Mapping (SLAM) solution. We use a pose graph-based approach to project the $3$D point clouds from the robot's LiDAR into a $2$D occupancy grid \cite{Rogers2014}. The robot tracks its \textit{poses}, i.e., locations and orientations along a trajectory, and integrates data from its IMU in order to localize and build a map in real-time without any a-priori information. Our approach manages the growth of inevitable sensor drift and does not rely on an external reference frame like GPS. 

Equipped with the ability to build accurate maps and localize reliably, the robot can then execute higher-level, autonomous capabilities. First, the robot is capable of autonomous planning and navigation using the ROS-based architecture from our previous work \cite{Gregory2016}. Our software suite leverages a global motion planner, local planner, and local controller to plan and execute kinematically-feasible routes to a designated pose, which effectively enables obstacle-aware autonomous navigation. 

Additionally, we implemented autonomous exploration using an information-theoretic, frontier-based approach \cite{Fung2016}. Frontiers are defined as the boundary of known and unknown space in the robot's occupancy grid \cite{Yamauchi1997}. The robot analyzes the current state of its map, designates frontiers, and then evaluates the utility of navigating to each frontier. As with our previous work, we define the frontier utility as the amount of expected information to be gained minus the amount of effort that is required. Information gain is measured as a function of new map cells for which LiDAR data will probabilistically be collected and effort is quantified by the length of the planned path from the robot's current position to the frontier. In effect, our approach to exploration seeks to minimize the entropy of the robot's graph by greedily navigating to places with the most unknown space. To help expedite the exploration process and provide the human with the ability to direct where they would like the robot to collect data, we have defined a keep-in region in our exploration implementation. Frontiers can only be created and selected if they are contained completely within the bounds of the keep-in area. For simplicity, this region is defined as a circle, parameterized by a center location and a radius; however, our system supports the definition of any convex polygon for the keep-in region. It is also important to note that all of our autonomous capabilities are platform agnostic and require a minimum set of common sensors consisting of a LiDAR and IMU. No camera data, GPS, communications, off-board processing, or post processing are required for the autonomous behaviors described in this section.

\section{Outfitting the Human Teammate}
\label{sec:human}
In the context of human-robot teaming in this work, the human is viewed as the high-level decision-maker to progress the specific mission-of-interest. This is because the human is able to incorporate mission requirements, context, intent, and preferences. As a result, the ability to task the robotic teammate intuitively, quickly, and accurately is crucial to accomplishing the mission efficiently.

\subsection{Hardware}
\label{subsec:human-hardware}
\begin{figure}[htbp]
  \centering
  \subfloat[Microsoft HoloLens Augmented Reality Head-Mounted Device (AR-HMD) \label{fig:HoloLens}]{%
    \includegraphics[width=0.48\linewidth]{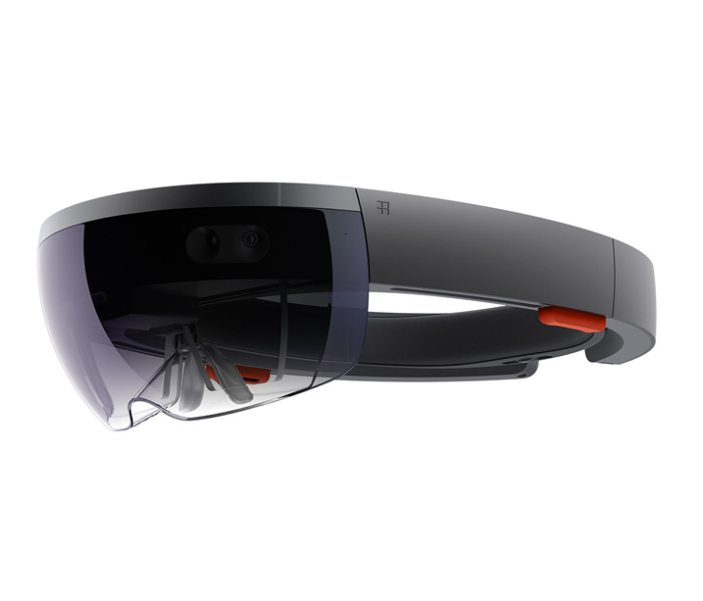}}
    \hfill
  \subfloat[Manus VR gesture glove \label{fig:glove}]{%
    \includegraphics[width=0.48\linewidth]{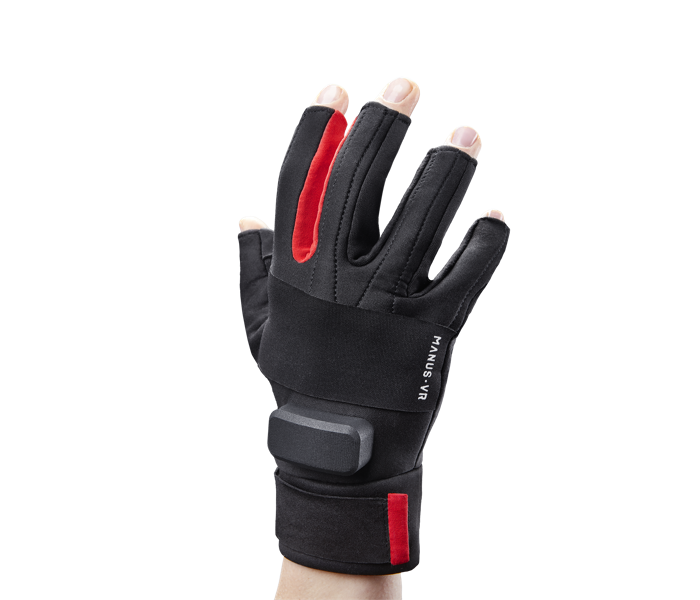}}
  \caption{The hardware used to outfit the human teammate and enable intuitive gesture command with enhanced situational awareness.}
  \label{fig:human} 
\end{figure}

The human, without additional equipment, cannot issue commands that are optimized for teaming in unstructured environments. One approach to sending commands could be for the robot to sense the human using imagery or LiDAR data; however, this requires constant line-of-sight operations, which is inefficient and impractical in many large, real-world environments. Instead, we take a distributed, sensors-based approach to issuing commands by outfitting the human with an AR-HMD and a gesture glove. We use the Microsoft HoloLens\nocite{Microsoft} AR device to overlay information collected by the robot into the view of the human in order to maximize situational awareness and aggregate the knowledge of both agents. The HoloLens device builds a local map of its surrounding environment, which we use to align and localize with the robot's map. The critical challenge, but necessary component, to enable a human wearing an AR-HMD and a robotic teammate to accurately share map data is online, mutual alignment of coordinate frames.  We leverage advancements from our previous work to do this in a robust way \cite{Reardon2018}. We also make use of the Manus VR gesture glove\nocite{ManusVR}, which uses built-in resistive sensors to track individual finger movement as well as an IMU to improve the accuracy of state estimation. Whilst the gesture glove uses a Bluetooth radio to transmit data to a receiver dongle, the driver software requires Microsoft Windows for operation. As such, our system utilised an Intel Compute Stick\nocite{IntelStick} which then communicated an interpreted signal to the robot over a $5$GHz radio via a TCP socket. Future development will improve the ability to handle the signal at a lower level thereby improving battery and CPU usage and increasing the portability across robotic platforms. The AR device and gesture glove that was used to outfit the human are shown in Figure~\ref{fig:human}. 

\subsection{Intuitive Commands}
\label{subsec:human-commands}
\begin{figure}[htbp]
  \centering
  \subfloat[The \textit{traverse} command, with three options for designating the distance of the destination relative to the human. \label{fig:goto}]{%
    \includegraphics[width=0.99\linewidth]{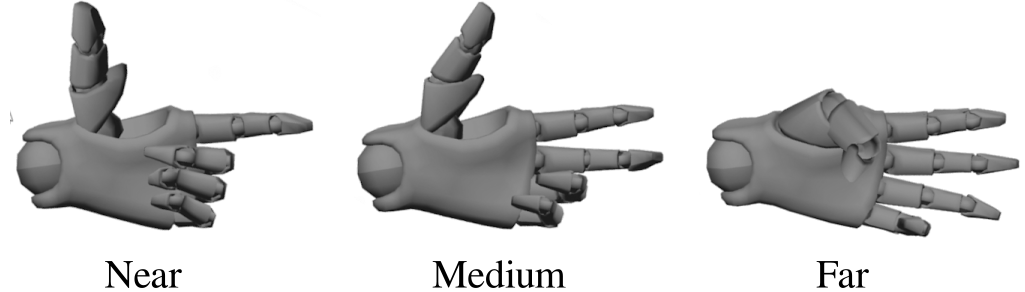}}
  \\
  \subfloat[The \textit{explore} command, with three options for designating the distance of the keep-in region, issued by moving the ``pointing" fingers of the traverse command. \label{fig:explore}]{%
    \includegraphics[width=0.99\linewidth]{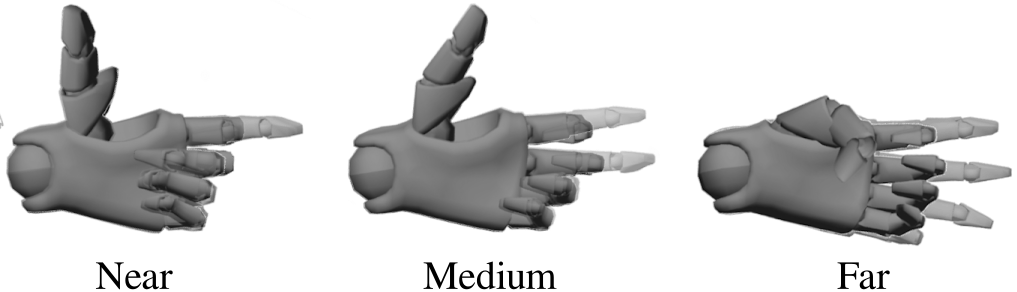}}
  \\
  \begin{minipage}{1.25in}
    \subfloat[The \textit{stop} command. \label{fig:stop}]{%
      \includegraphics[height=0.8in]{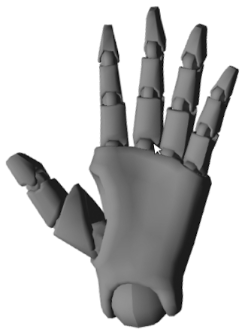}}  
  \end{minipage}
  \begin{minipage}{1.25in}
    \subfloat[The \textit{return} \newline command. \label{fig:return}]{%
      \includegraphics[height=0.7in]{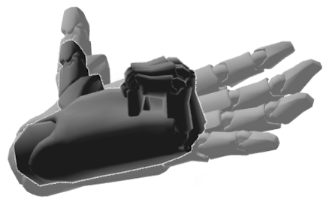} } 
  \end{minipage}
  \caption{The proposed command set and corresponding gestures used in our system.} 
  \label{fig:gestures} 
\end{figure}

Once outfitted with a gesture glove, the human's finger and hand movements can be tracked; however, there is still a paramount design decision for realizing an effective human-robot team and that is: what commands are necessary and what gestures should be used for each command? Commands need to be sufficiently different so that they can be interpreted quickly and accurately, but a large set of complex commands can become burdensome for the human to remember or require more time to train and learn. We do not seek to derive the optimal set of commands or gestures for all missions, but rather present a feasible solution and evaluate its usefulness as a first step towards understanding the requirements for optimal gesture control. Here, we propose a set of four commands that provide sufficient, foundational control and, for each, describe our gesture design, as shown in Figure~\ref{fig:gestures}. 

To begin, we define a gesture for indicating that a command is to be issued. Since the human will naturally use their hands through out a typical mission to do things other than sending commands, it is imperative that the gesture controller only issue commands when the human intends on it. Otherwise, every hand movement could be misinterpreted and commands would be sent erroneously. The human indicates to the gesture glove that the next sequence of finger and hand movements correspond to a command by making a fist for approximately $500$ms. We refer to this as \textit{activation} and provide vibro-haptic feedback to the user through a short vibration in the glove to notify the human that a command is ready to be received. In this case, the vibration pattern was a single quick pulse. Once the glove vibrates, the human is able to do a hand gesture corresponding to one in our defined set and issue a command of their choosing. Should the gesture not be recognised the system is able to notify the operator by providing a single long pulse, indicating a failed recognition of the intended gesture.

1) \textit{Traverse Command}: An elementary command for tasking a robot is autonomous navigation control, referred to here as \textit{traverse} or ``goto.'' This command allows the human to task its teammate to navigate from the robot's current location to a designated goal location, e.g., path planning and autonomous navigation around obstacles. This provides the human with a method by which the robot can rapidly be instructed where to move, but balances workload because the human is not required to manually drive the robot to the goal. Furthermore, the traverse command is the basis for many higher-level applications because so many capabilities require autonomous navigation to a specific location. For the traverse command, we chose a natural gesture that captures the essence of fast, point-and-go tasking, and take inspiration from tactical hand signaling used in the military. We chose to use the gesture of \textit{pointing} to issue a traverse command and used the direction of the human's gaze, as measured by the AR-HMD, to issue the location and orientation of the goal destination. We increased the expressivity of the traverse command by providing the human with three distances for which the robot could be task. These were \textit{near}, \textit{medium}, and \textit{far}, corresponding to $2.0$, $4.5$, and $7.0$m from the human's current position. To designate one of these three options, the human used one, two, or three fingers, respectively. Importantly, this value could be set in the field to suit the environmental context. For example, a far location may extend to $1000$m in an open field, yet only $10$m in an indoor environment.

2) \textit{Explore Command}: A more sophisticated command that a human would likely require in a real-world mission involving human-robot teams is the ability to instruct the robot to \textit{explore} a region of the environment. This removes the burden of point-to-point control and allows the human to multi-task because the robot is autonomously deciding where to navigate next. At its core, exploration is a series of traverse commands, except that the robot is intelligently assigning itself goal locations. This command leverages our information-theoretic exploration capabilities and synthesizes the human's desire for rapid data acquisition with the robot's autonomous behaviors. To issue this command, the human gestures in a similar fashion for the traverse command, but additionally moves their finger back and forth. The dynamics of the finger are registered using the gesture glove and an explore command is issued instead of a traverse command. As described in the Autonomous Behaviors Section, the robot's exploration capability is parameterized by a keep-in region. When the human issues an explore command, they define the location and size of the keep-in region using one dynamic finger for a near region $7.0$m away from the human with a radius of $7.0$m, two fingers for a medium region $15.0$m away from the human with a radius of $15.0$m, or three fingers for a far region $25.0$m away from the human with a radius of $25.0$m radius. Again, these parameters are easily reconfigurable to meet mission-specific needs. 


3) \textit{Stop Command}: A crucial command for human-robot teaming is the \textit{stop} command. As the mission progresses, the human and robot teammates collect information that may change the prioritization of tasks or location of operations. Similarly, the robot may progress into areas of lesser importance or need to be halted for safety concerns. We empower the human to issue a stop command which preempts the robot's current command, and stops navigation if the robot is moving during a previous traverse or explore command. The stop command is sent by the human using an outward-facing, open palm. 

4) \textit{Return Command}: The final command that could be necessary for intuitive human-robot teaming is the \textit{return} command. In a typical mission, the robot may operate at great distances from the human either through a series of traverse or exploration commands. While the human could potentially issue a series of traverse commands to instruct the robot to return, we have introduced this single, effortless command to assist the human. The robot uses knowledge of the human's pose localized in its own map and autonomously plans to a goal location $1$m in front of where the human is looking. This location serves as a goal for autonomous navigation, similar to a traverse command. 

\begin{figure}[htbp]
  \centering
  \includegraphics[width=0.60\linewidth]{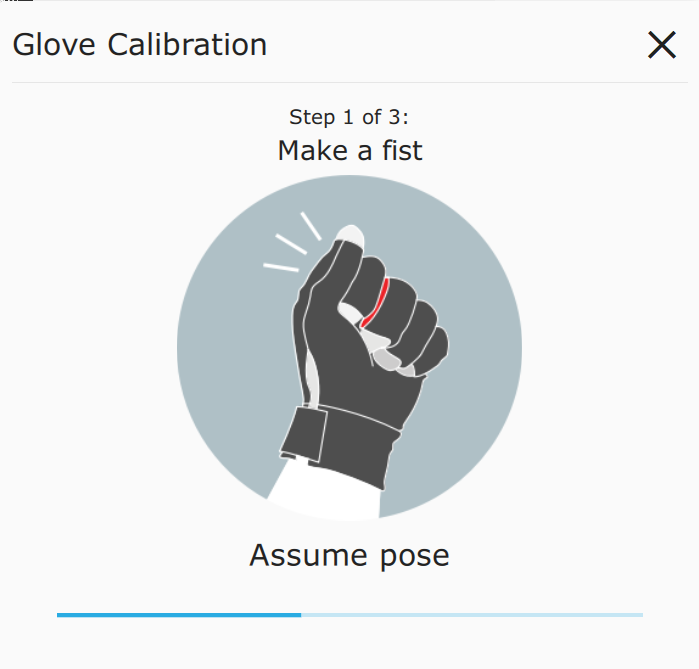}
  \caption{ManusVR calibration software provided a simple method for gesture calibration between participants.} 
  \label{fig:manusvr-calibration} 
\end{figure}

\textit{Calibration}: To maximise the likelihood of recognizing the intended gesture, we calibrated our system using the ManusVR glove software as shown in Figure \ref{fig:manusvr-calibration}. This consisted of a short sequence of gestures, which were shown to the human and asked to be copied.

\section{Field Experiment}
\label{sec:testing}
To test the usefulness of our proposed command set, gesture definitions, and autonomous behaviors supporting information exchange in human-robot teaming, we performed a field experiment. The details of the environment and preliminary, anecdotal results from human participants are included in this section.

\subsection{Pilot Study}
\label{subsec:preliminary}
\begin{figure}[htbp]
  \centering
  \subfloat[Environment $1$ \label{fig:env-1}]{%
        \includegraphics[width=0.32\linewidth]{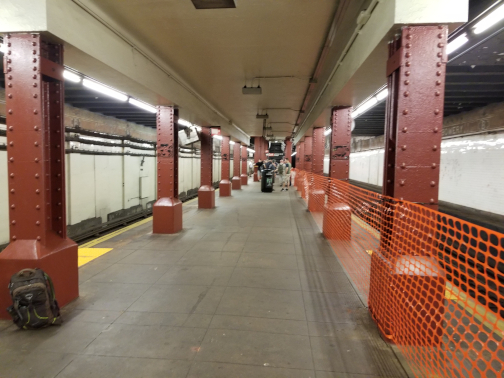}}
    \hfill
  \subfloat[Environment $2$ \label{fig:env-2}]{%
        \includegraphics[width=0.32\linewidth]{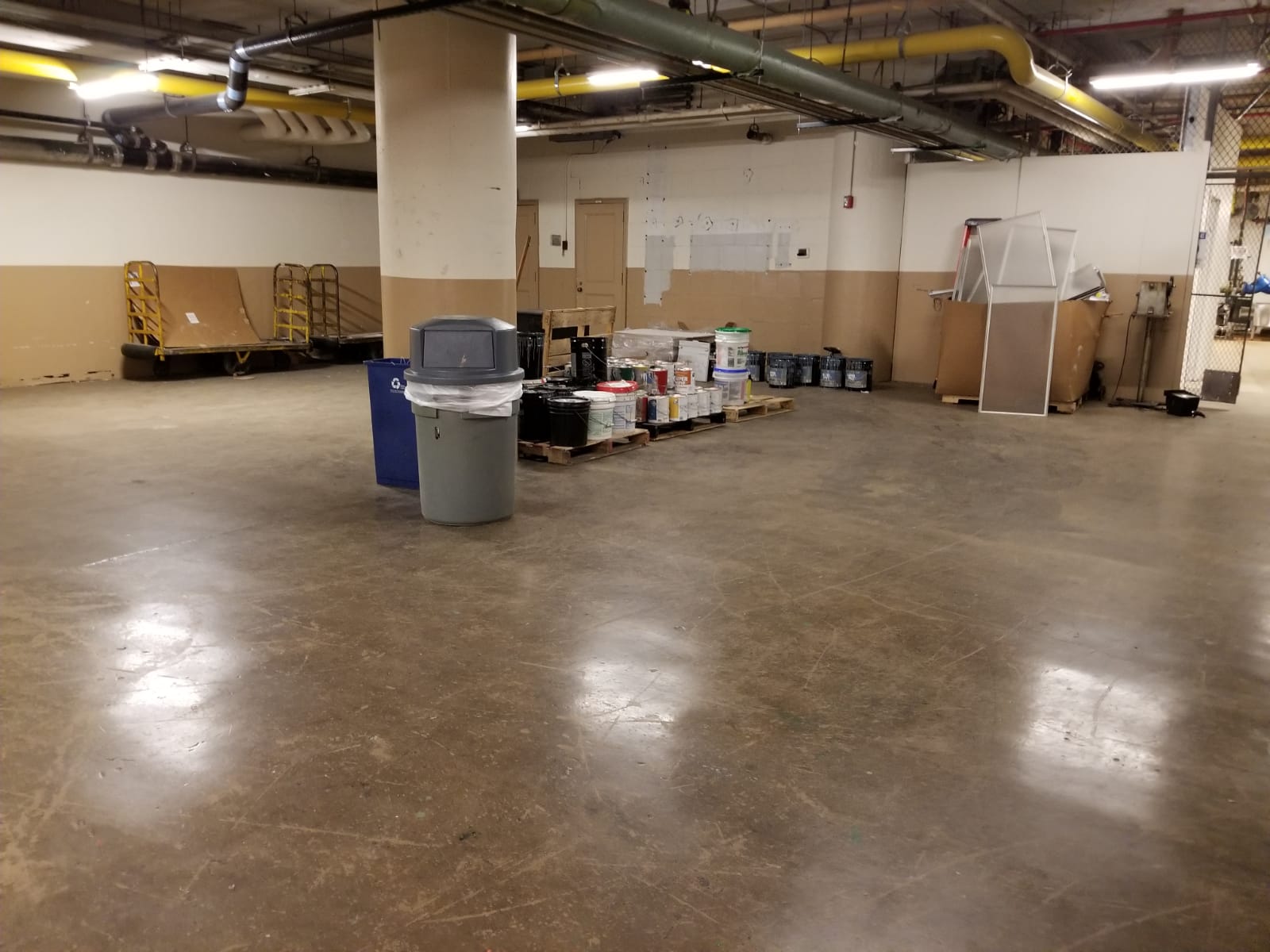}}
    \hfill
  \subfloat[Environment $3$ \label{fig:env-3}]{%
        \includegraphics[width=0.32\linewidth]{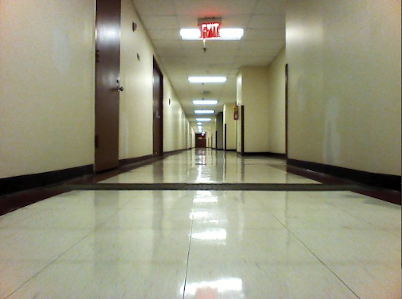}}
  \caption{The three operational environments in which the human-robot teaming system pilot study was conducted.}
  \label{fig:environments} 
\end{figure}

To evaluate applicability and excite the potential failure modes of our solution,
we tested our AR-enabled, gesture control system in three separate environments. 
The first environment was a dirty subway station platform with challenging, 
real-world navigation constraints due to the narrow space. 
The second environment was in the basement of an office building with intermittent lighting and considerable clutter that also complicated autonomous navigation. 
Finally, the third environment was the upper floor of an office building that had better lighting but was more austere with fewer features, which introduces challenges in feature-based mapping and localization. 
Representative photos from each of the three environments are shown in Figure~\ref{fig:environments}. 

In each of these three environments, we asked six participants, with no prior experience in the environment or with our system, to task the robotic teammate and produce a map as quickly as possible. The context of this pilot study was to emulate a reconnaissance-like mission in a real-world environment in an effort to better understand the performance and usefulness of our system. After each trial, the participants were asked to describe their experience and overall opinion of the system as it relates to usability and function.

\subsection{Preliminary Results}
\label{subsec:results}
Results from the field experiment
indicate that a human possessing no prior experience with our system can successfully command a robotic teammate in an unknown environment, without external instrumentation. 
In all six trials, for each of the three environments, the humans were able to direct the robot to 
build a map using our defined gestures and the visualized structure of the environment from our augmented reality display. 
This suggests that our proposed set of commands are adequate for human-to-robot communication and enables some amount of human-robot teaming. 
It also leads us to believe that gestures assigned to each command are sufficiently repeatable for use. 
Furthermore, we note several topics of consistency and robustness for gesture control in the Lessons Learned section. 

\begin{figure}[htbp]
  \centering
  \subfloat[LoS command \label{fig:los-command}]{%
        \includegraphics[width=0.43\linewidth]{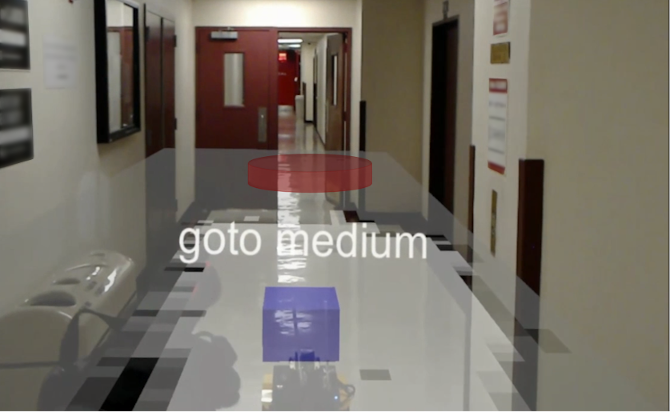}}
    \\
  \subfloat[Robot's plan \label{fig:los-plan}]{%
        \includegraphics[width=0.50\linewidth]{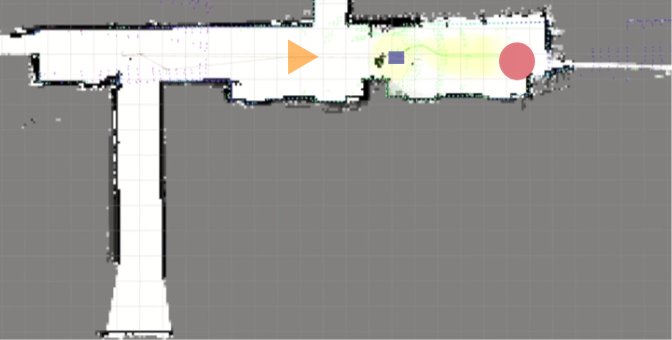}}
    \hfill
  \subfloat[Execution visualization \label{fig:los-execution}]{%
        \includegraphics[width=0.43\linewidth]{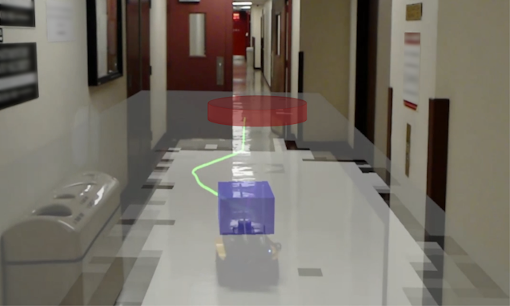}}
  \caption{An example of a \textit{Line-of-Sight} operations. (a) The human views the robot and map through the HoloLens and issues a \textit{traverse medium} gesture command (annotated ``goto medium'' in the image). (b) Then, the robot plans a path (green line) to the designated goal location (red disc) relative to the human (orange triangle), as shown in this top-down, orthogonal occupancy map. (c) Finally, the human visualizes the goal location, robot's planned path, and navigation in the AR-HMD display. }
  \label{fig:los} 
\end{figure}

In addition to supporting our hypothesis that AR and gesture control, together, support the efficiency of data flow in human-robot teams, we also observed several concrete examples of useful functionality offered by our system. The first natural use case that arose during our pilot study was improved \textit{Line-of-Sight} (LoS) operations. Rather than using a wireless joystick, or mouse and computer monitor, to task the robot, the participants sent many traverse commands and walked behind the robot as it autonomously navigated, as illustrated in Figure~\ref{fig:los}.

\begin{figure}[htbp]
  \centering
  \subfloat[NLoS command \label{fig:nlos-command}]{%
        \includegraphics[width=0.43\linewidth]{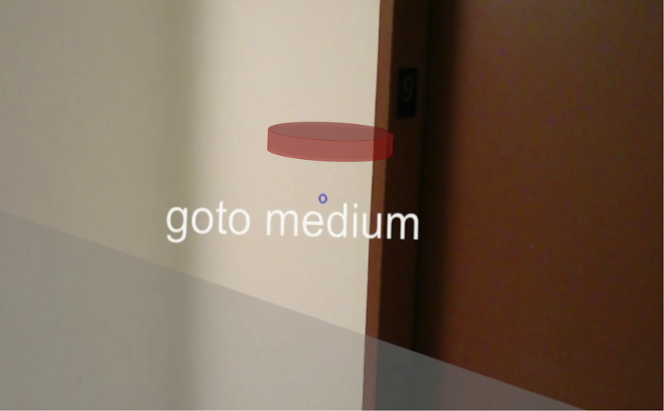}}
    \\
  \subfloat[Robot's plan \label{fig:nlos-plan}]{%
        \includegraphics[width=0.50\linewidth]{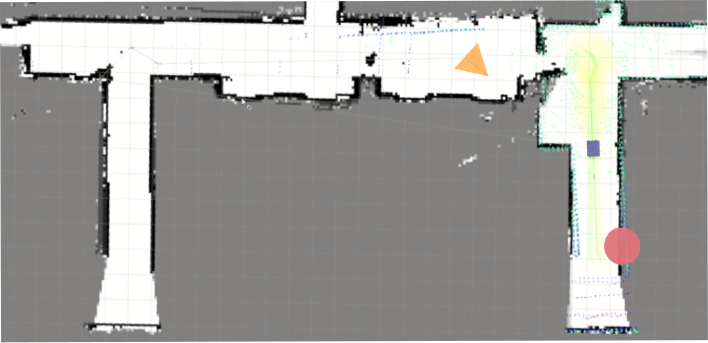}}
    \hfill
  \subfloat[Execution visualization \label{fig:nlos-execution}]{%
        \includegraphics[width=0.41\linewidth]{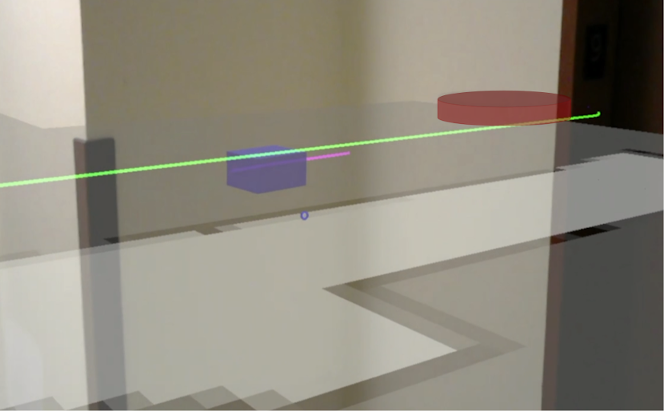}}
  \caption{An example of a \textit{Non-Line-of-Sight} operations where a human sends a command for the robot to navigate to a location on the other side of the wall from the human. Enhanced situational awareness is provided by our system using a blue box for the robot's location to assist in understanding where the robotic teammate is and how the map is being generated.} 
  \label{fig:nlos} 
\end{figure}

The second useful mode of operation observed during our pilot study was \textit{Non-Line-of-Sight} (NLoS). In this case, the human could not physically see the robot because there was a wall or obstacle between the two teammates. However, by displaying a blue box at the robot's location in the human's AR-HMD display, our system uniquely offers the capability to observe the virtual location of a teammate and enables continued tasking in NLoS situations. This greatly enhances situational awareness because the participant knows where its teammate is at all times, regardless of distance or environmental obstructions. An example of a participant issuing a command to the robot, through a wall, to navigate in a portion of the environment the human has no knowledge of is shown in Figure~\ref{fig:nlos}.

\begin{figure}[htbp]
  \centering
  \subfloat[Object-of-interest in camera view \label{fig:obj-det-camera}]{%
        \includegraphics[width=0.38\linewidth]{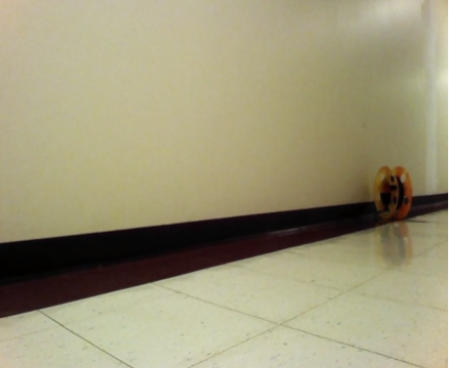}}
    \\
  \subfloat[Object localized in robot's map \label{fig:obj-det-robot}]{%
        \includegraphics[width=0.50\linewidth]{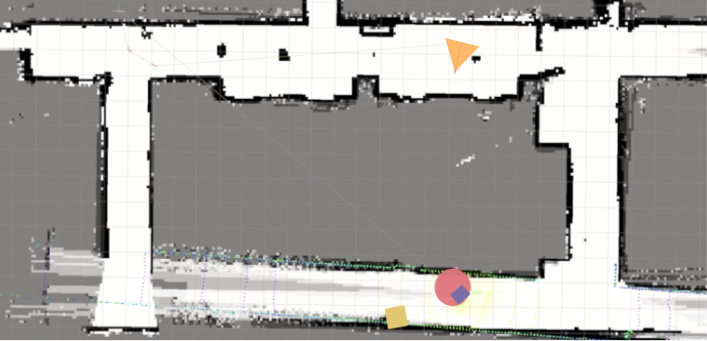}}
    \hfill
  \subfloat[Object visualization \label{fig:obj-det-viz}]{%
        \includegraphics[width=0.41\linewidth]{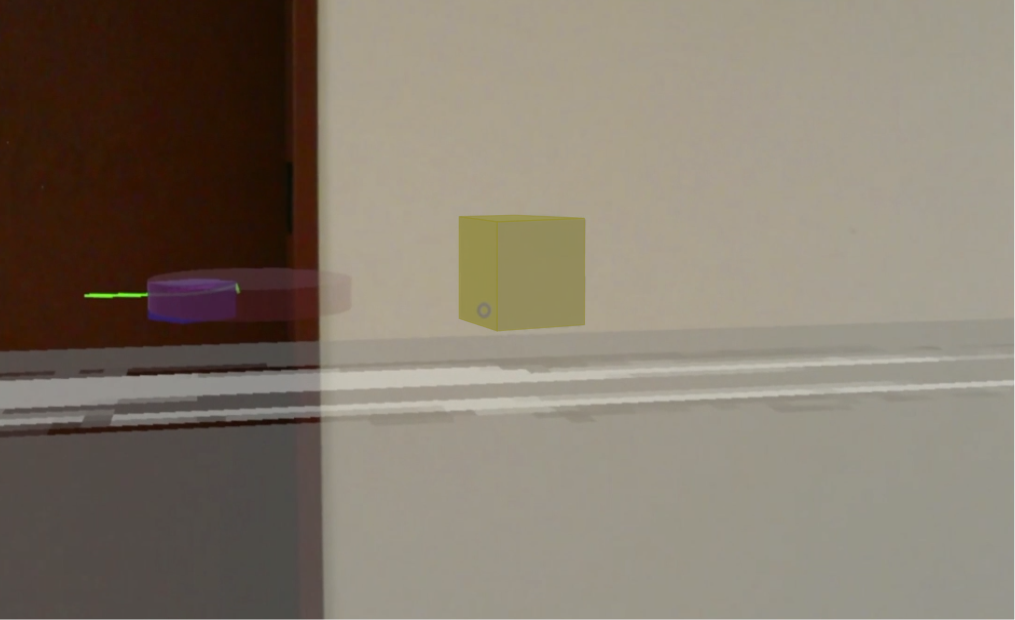}}
  \caption{An example of object-of-interest (yellow box) inspection by way of detection, localization, and visualization.} 
  \label{fig:obj-det} 
\end{figure}

While robust autonomous object detection is outside the scope of this work, we found that our system supports enhanced situational awareness through the visualization of objects-of-interest and areas-of-interest. 
We have specifically designed our system architecture such that visualization markers can be manually or autonomously inserted into both the robot's map as well as the AR-HMD display so that the teammates can recognize, localize and leverage new information. An example of this is shown in Figure~\ref{fig:obj-det} where an object-of-interest was manually added in the robot's map once it became in field-of-view of the robot's onboard camera. The robot then shared this information with the human by placing a cube in the AR-HMD display at the corresponding location, which in some cases prompted new or additional tasks by the human for further investigation. 

An unexpected, but noteworthy method of control that emerged during our pilot study is fast agent re-positioning through command sequencing. After a series of mission-required traverse and exploration commands, the robot often times navigated a considerable distance away from the human while collecting data. Rather than tediously issuing several traverse commands to move the robot, we observed that participants would send a return command and then, after allowing the robot to navigate for some time, a subsequent stop command once the robot was in position for the next task in the mission. The human did not actually require the robot to return, and in fact stopped navigation before the robot completed its planned navigation, but rather used this command sequence to assist in the efficient movement of its teammate. By pairing the two commands, the human could have the robot plan and navigate over longer distances relative to traverse commands while issuing fewer commands and requiring less oversight of navigation goal locations. 
\begin{figure}[htbp]
  \centering
  \subfloat[A return command that initiates longer-distance navigation \label{fig:return-command}]{%
        \includegraphics[width=0.43\linewidth]{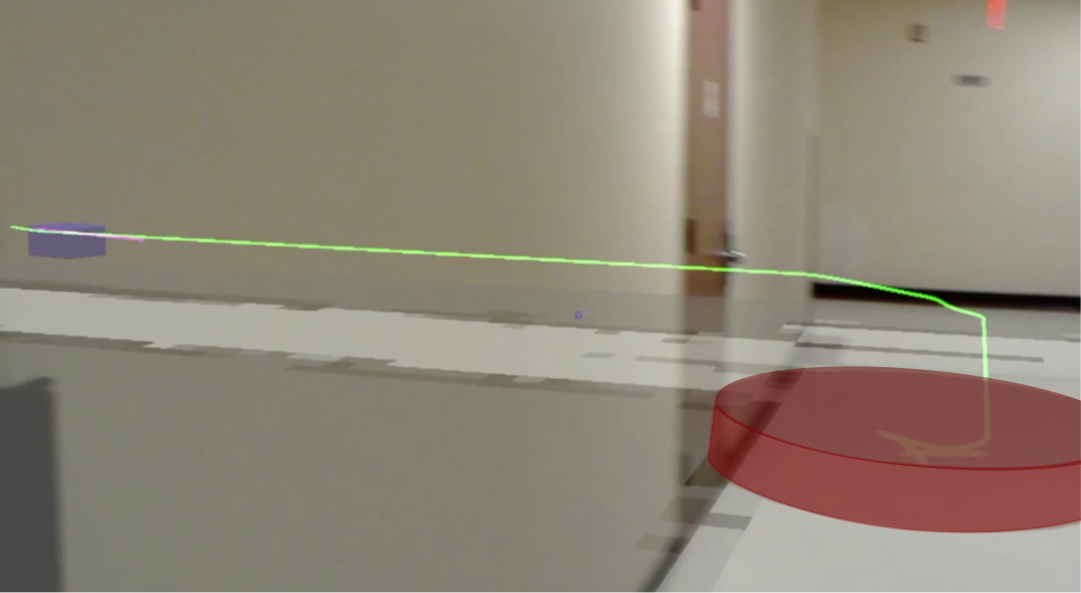}}
    \\
  \subfloat[Robot navigating along planned path \label{fig:return-plan}]{%
        \includegraphics[width=0.50\linewidth]{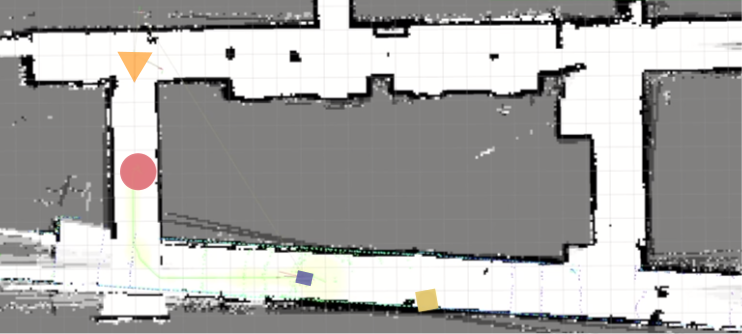}}
    \hfill
  \subfloat[A preemptive stop command once the robot is at the hallway intersection \label{fig:stop-command}]{%
        \includegraphics[width=0.43\linewidth]{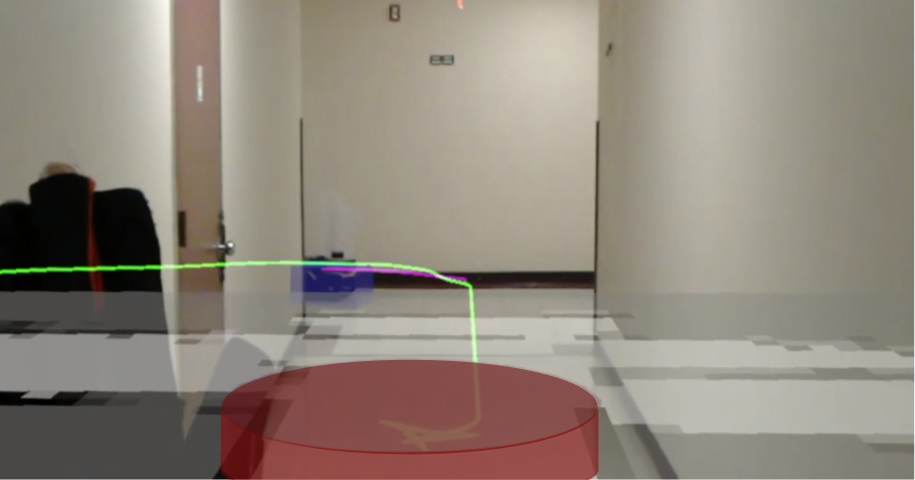}}
  \caption{An example of the human quickly and effortlessly re-positioning the robot using a series of \textit{return} and \textit{stop} commands.} 
  \label{fig:reposition} 
\end{figure}

The final example of useful functionality that we observed our system provides was exploration of an unknown environment from greater standoff distances. Once the participant felt comfortable that the robot was able to autonomously navigate, they often times sent an exploration command with a medium to large keep-in region. As expected, exploration alleviated the participant from having to decide where the robot should autonomously navigate to. Because the humans typically issued exploration commands in parts of the environment that they could not physically see for themselves, this mode of operation inherently allowed for operation at greater standoff distances. An example of one of the exploration instances during our pilot study is shown in Figure~\ref{fig:explore}.
\begin{figure}[htbp]
  \centering
  \subfloat[Robot's evaluation \label{fig:explore-plan}]{%
        \includegraphics[width=0.50\linewidth]{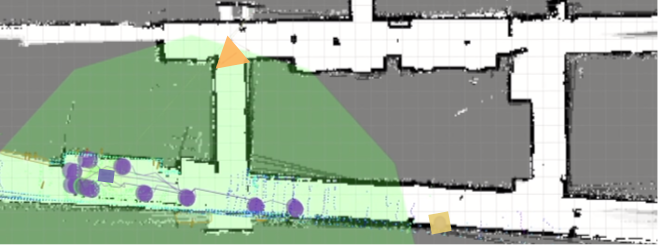}}
    \hfill
  \subfloat[Exploration visualization \label{fig:explore-execution}]{%
        \includegraphics[width=0.39\linewidth]{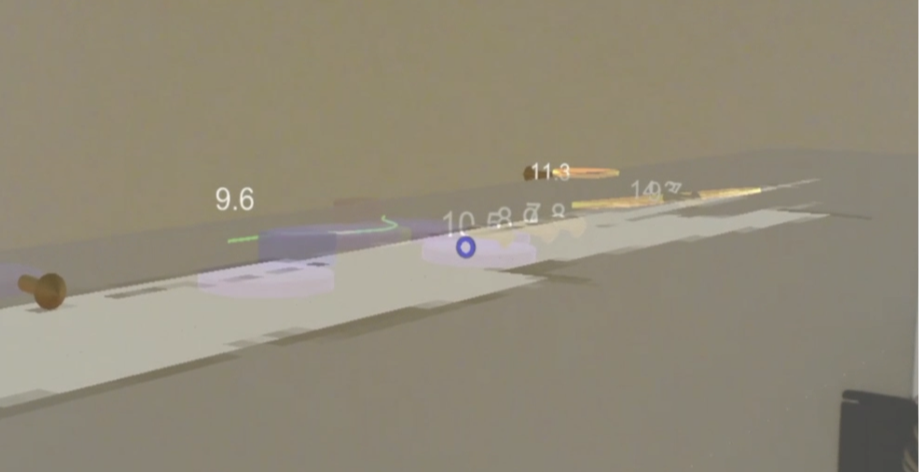}}
  \caption{An example of the robot exploring within a keep-in region (green disc) while the human is at an increased standoff distance.} 
  \label{fig:explore} 
\end{figure}

\section{Lessons Learned}
\label{sec:lessons}
Our pilot study provided preliminary results that revealed strengths and shortcomings of our proposed system, which we intend to use to improve the next iteration of system design. 
In general, we gained insight into the the robot, the HoloLens and our AR display, the glove, our proposed command set and corresponding gestures, and the complex interactions between the components of this system.
We present these learned lessons, and identify open research questions, for the interest of both practitioners and researchers alike. 

\emph{Challenges of robotic hardware, mapping, and navigation in subterranean and confined environments:}
With regards to the robot, we learned that overheating is a genuine concern that must be accounted for when designing and deploying systems in real-world environments. We observed hardware malfunction and degradation with the platform, sensors, and processing payloads in the unconditioned, subway environment. 
While overheating concerns are not new to system engineering, it was still surprising how impactful the weather and heat were in an environment that is not conventionally thought of as having overheating issues, e.g., subterranean. 
Additionally, we observed that SLAM solutions and motion planners capable of handling dynamic obstacles are crucial in human-robot teaming because humans are inherently dynamic obstacles working in close proximity to the robot. 
Without proper handling of dynamic obstacles in the mapping, planning and navigation architecture of the robot's autonomous behaviors, the human teammate could actually hinder the robot's capabilities and negate the potential benefit of a multi-agent team.

\emph{Overcoming the limitations of mixed reality devices:}
Similar to the robot, the Microsoft HoloLens also appeared to experience considerable overheating in the demanding operational environment. 
One workaround we found to be successful was to make use of a second, fully-charged HoloLens that was originally intended to be a backup.  
Upon overheating of one device we could fail-over to the second while the first device cooled down; 
although, a preferred solution would be an AR device that is specifically engineered for the extreme weather elements. 
Additionally, we confirmed and discovered several factors for improving performance during our pilot study. 
First, we found that displaying a 3D icon (blue box) over the robot in the AR-HMD display enhances situational awareness 
by providing a reference for the robot's position as it operates beyond line of sight collecting map data. 
Likewise, displaying text corresponding to which command was sent to the robot was the preferred method of non-intrusive feedback for confirming the human's intent. 
We hypothesize that a similar textual display for object detection will greatly enhance situational awareness, especially when a robot perceives an object that the human never physically observes, e.g., NLoS operations. 
The AR display opacity, map height, and amount of information displayed 
all affect both user experience and performance, 
and therefore should all be easily customizable. 
We also found considerable improvement in both the SLAM performance and map visualization by providing the robot and AR-HMD multiple, initial transformations for the relative offset between the two devices at the start of every trial. 
Related to this, the HoloLens also appears to have similar drift concerns as the robot and gesture glove, all of which are IMU-based state estimation. 
Considerable research, development, and engineering efforts of software and hardware solutions are required for a fieldable system that is capable of operating over long distances, durations, and extreme movement. 
The HoloLens used here appears to build a map where the structure is not updated at a frequency that supports real-time operations in dynamic, unstructured environments. 
This became an issue when other slow-moving non-participants inadvertently became part of the HoloLens map.
A fieldable solution will likely require constant scanning and updating of the mesh generation and map composition. Many current commercial AR devices may employ single-scan approaches because they assume typical use will be gaming and entertainment applications in relatively-static environments; however, this is insufficient in real-world missions. 

\emph{Providing consistent and robust gesture control:}
We learned several lessons with regards to the glove as it relates to gesture control. First, glove calibration should be performed for each individual user because humans have subtle differences in finger placement for gestures. 
On our device this is especially true for thumb placement, e.g., making a fist and placing the thumb on top or on the side of the fist. 
Furthermore, calibration should be completed in a natural, but controlled manner because humans may be inconsistent when it comes to issuing commands relative to the gestures they used during calibration. 
A human may become lazy or fatigued over the course of a mission such that their gestures differ considerably from the more deliberate gestures they used during calibration. 
This lesson also motivates the development of robust algorithms for gesture control so that considerable variations in gesture detection for a user can be handled appropriately. 
To this end, we noticed that the gesture device often times accumulated an irrecoverable amount of drift in its state estimation and required re-calibration after extended periods of time. 
This lesson suggests there is a need for the advancement of algorithms that bound and remove sensor drift from gesture devices and that standard operating procedure for a robust fieldable system might consider a protocol to re-calibrate and verify glove signals often. 
A fieldable system should also optimize the communication hardware and protocol to ensure that even if the robot and human do not remain in constant communications range for transferring most data, 
the human still has sufficient bandwidth to send commands so that control is never lost.

\emph{Expressivity, flexibility, and precision in robot command:}
The command set and associated gestures appeared to provide sufficient expressivity and usefulness, but there is room for improvement. Participants reported a desire for a \textit{Follow Me} command that requires no goal location, but allows the human to walk naturally to a new location while the robot autonomously navigates behind them. The appropriate distance and velocity that the robot maintains 
in the presence of obstacles and environment structure 
is an open research question and is likely mission-dependent. 
In terms of our implemented commands and gestures, the activation command and vibration feedback were universally appreciated because of the efficient confirmation of command readiness. 
On the other hand, the hard-coded distances were simplistic and an insufficient means for commanding robots with fine-grained control.
Participants sometimes found it challenging to send commands with the appropriate distance in NLoS operations because the goal location is relative to the human's position and they cannot physically see the environment where the goal could be placed. 
A more robust specification of goal locations is required to maximize usability and intuitive control. 
Along these lines, we found that participants required more fine-tuned, flexible control with respect to the orientation of the robot's final pose once at a navigation goal in the event the human wanted the robot to view a particular object or direction. 
Lastly, we discovered that not all participants possessed equal dexterity in moving individual fingers while keeping other fingers in place. This is an important consideration for system designers defining gestures because it can pose a potential issue for dynamic gestures, especially if commands differ by a small amount or require exact finger placement.

\emph{Usefulness of NLoS operation:}
While NLoS operation was a planned part of our system's function,
none of the authors were prepared for how intuitively impactful the experience would be for the users.
Anecdotally, when the participants were instructed to begin the NLoS operation of the robot for the first time,
all or nearly all verbalized to indicate the impressiveness of the capability, even at this preliminary stage.
It therefore bears remarking that
the ability to direct an autonomous asset that is beyond a human's line of sight, and visualize its 
sensing and planning situated in one's own reality,
whether for saving the human time or keeping him/her out of harm's way,
is one of the easiest-to-convey impacts for a variety of real-world domains.


\section{Conclusions}
\label{sec:conclusions}
In this work, we presented and analyzed an AR-enabled, gesture-based system for improved communications in a human-robot team. Our proposed system shows promise of real-world applicability for service-oriented missions, and the lessons learned from our experimental field study may help direct future development, fielding, and experimentation of autonomous HRI systems. From our preliminary results, we have identified ways in which our system could be improved. First, sensor drift and its impact on state estimation are a paramount concern for multi-agent mapping. Future efforts will include improved techniques for managing the inevitable sensor errors that accumulate over time, especially considering our system makes use of IMUs on the robotic teammate, AR-HMD, and gesture glove. Similarly, our approach for frame alignment between the AR-HMD and robot maps should be optimized to enable long-duration missions with complex movement in unstructured terrain. 

\bibliographystyle{aaai}
\bibliography{references}

\begin{thebibliography}{}

\bibitem[\protect\citeauthoryear{ast}{}]{astra}
{Orbbec Astra Series}.
\newblock \url{https://orbbec3d.com/product-astra-pro/}.
\newblock Accessed: July 25, 2019.

\bibitem[\protect\citeauthoryear{Bul}{}]{Bullet}
{Bullet M Titanium Datasheet}.
\newblock \url{https://www.ui.com/airmax/bulletm/}.
\newblock Accessed: July 25, 2019.

\bibitem[\protect\citeauthoryear{Cacace, Finzi, and
  Lippiello}{2016}]{Cacace2016}
Cacace, J.; Finzi, A.; and Lippiello, V.
\newblock 2016.
\newblock {Multimodal interaction with multiple co-located drones in search and
  rescue missions}.
\newblock {\em arXiv preprint arXiv:1605.07316}.

\bibitem[\protect\citeauthoryear{Elliott, Hill, and Barnes}{2016}]{Elliott2016}
Elliott, L.~R.; Hill, S.~G.; and Barnes, M.
\newblock 2016.
\newblock {Gesture-based controls for robots: overview and implications for use
  by Soldiers}.
\newblock Technical report, US Army Research Laboratory Aberdeen Proving Ground
  United States.

\bibitem[\protect\citeauthoryear{Fung \bgroup et al\mbox.\egroup
  }{2016}]{Fung2016}
Fung, N.~C.; Nieto-Granda, C.; Gregory, J.~M.; and Rogers, J.~G.
\newblock 2016.
\newblock {Autonomous Exploration Using an Information Gain Metric}.
\newblock Technical report, US Army Research Laboratory Adelphi United States.

\bibitem[\protect\citeauthoryear{Gadre \bgroup et al\mbox.\egroup
  }{2019}]{Gadre2019}
Gadre, S.~Y.; Rosen, E.; Chien, G.; Phillips, E.; Tellex, S.; and Konidaris, G.
\newblock 2019.
\newblock End-user robot programming using mixed reality.
\newblock In {\em Proceedings of the IEEE International Conference on Robotics
  and Automation (in press). IEEE}.

\bibitem[\protect\citeauthoryear{Gregory \bgroup et al\mbox.\egroup
  }{2016}]{Gregory2016}
Gregory, J.; Fink, J.; Stump, E.; Twigg, J.; Rogers, J.; Baran, D.; Fung, N.;
  and Young, S.
\newblock 2016.
\newblock {Application of Multi-Robot Systems to Disaster-Relief Scenarios with
  Limited Communication}.
\newblock In {\em Field and Service Robotics},  639--653.
\newblock Springer.

\bibitem[\protect\citeauthoryear{Huang \bgroup et al\mbox.\egroup
  }{2019}]{Huang2019}
Huang, B.; Bayazit, D.; Ullman, D.; Gopalan, N.; and Tellex, S.
\newblock 2019.
\newblock {Flight, Camera, Action! Using Natural Language and Mixed Reality to
  Control a Drone}.
\newblock In {\em IEEE International Conference on Robotics and Automation
  (ICRA)}.

\bibitem[\protect\citeauthoryear{Int}{}]{IntelStick}
{Intel Compute Stick}.
\newblock
  \url{https://www.intel.com/content/www/us/en/products/boards-kits/compute-stick.html}.
\newblock Accessed: July 25, 2019.

\bibitem[\protect\citeauthoryear{Jac}{}]{Jackal}
{Jackal Unmanned Ground Vehicle}.
\newblock
  \url{https://clearpathrobotics.com/jackal-small-unmanned-ground-vehicle/}.
\newblock Accessed: July 25, 2019.

\bibitem[\protect\citeauthoryear{Man}{}]{ManusVR}
{Manus VR Gesture Gloves}.
\newblock \url{https://manus-vr.com/}.
\newblock Accessed: July 25, 2019.

\bibitem[\protect\citeauthoryear{Mic}{a}]{Microstrain}
{Lord MicroStrain Sensing 3DM-GX5-25 AHRS}.
\newblock \url{https://www.microstrain.com/inertial/3dm-gx5-25}.
\newblock Accessed: July 25, 2019.

\bibitem[\protect\citeauthoryear{Mic}{b}]{Microsoft}
{Microsoft Hololens}.
\newblock \url{https://www.microsoft.com/en-us/hololens}.
\newblock Accessed: July 25, 2019.

\bibitem[\protect\citeauthoryear{Quigley \bgroup et al\mbox.\egroup
  }{}]{Quigley2009ros}
Quigley, M.; Faust, J.; Foote, T.; and Leibs, J.
\newblock {ROS: an open-source Robot Operating System}.

\bibitem[\protect\citeauthoryear{Reardon \bgroup et al\mbox.\egroup
  }{2019}]{Reardon2019}
Reardon, C.; Lee, K.; Rogers, J.~G.; and Fink, J.
\newblock 2019.
\newblock {Augmented Reality for Human-Robot Teaming in Field Environments}.
\newblock In {\em International Conference on Human-Computer Interaction},
  79--92.
\newblock Springer.

\bibitem[\protect\citeauthoryear{Reardon, Lee, and Fink}{2018}]{Reardon2018}
Reardon, C.; Lee, K.; and Fink, J.
\newblock 2018.
\newblock {Come see this! Augmented reality to enable human-robot cooperative
  search}.
\newblock In {\em 2018 IEEE International Symposium on Safety, Security, and
  Rescue Robotics (SSRR)},  1--7.
\newblock IEEE.

\bibitem[\protect\citeauthoryear{Rogers, Fink, and Stump}{2014}]{Rogers2014}
Rogers, J.~G.; Fink, J.~R.; and Stump, E.~A.
\newblock 2014.
\newblock {Mapping with a ground robot in GPS denied and degraded
  environments}.
\newblock In {\em 2014 American Control Conference},  1880--1885.
\newblock IEEE.

\bibitem[\protect\citeauthoryear{Szafir}{2019}]{Szafir2019}
Szafir, D.
\newblock 2019.
\newblock {Mediating Human-Robot Interactions with Virtual, Augmented, and
  Mixed Reality}.
\newblock In {\em International Conference on Human-Computer Interaction},
  124--149.
\newblock Springer.

\bibitem[\protect\citeauthoryear{Vel}{}]{Velodyne}
{Velodyne Lidar Puck}.
\newblock \url{https://velodynelidar.com/vlp-16.html}.
\newblock Accessed: July 25, 2019.

\bibitem[\protect\citeauthoryear{Walker \bgroup et al\mbox.\egroup
  }{2018}]{Walker2018}
Walker, M.; Hedayati, H.; Lee, J.; and Szafir, D.
\newblock 2018.
\newblock Communicating robot motion intent with augmented reality.
\newblock In {\em Proceedings of the 2018 ACM/IEEE International Conference on
  Human-Robot Interaction},  316--324.
\newblock ACM.

\bibitem[\protect\citeauthoryear{White and Hill}{2018}]{White2018}
White, G., and Hill, S.~G.
\newblock 2018.
\newblock {Usability of a Gesture Control Device for Robotic Platform
  Operation}.
\newblock {\em International Conference on Science and Innovation for Land
  Power (ICSILP)}.

\bibitem[\protect\citeauthoryear{Whitney \bgroup et al\mbox.\egroup
  }{2018}]{Whitney2018}
Whitney, D.; Rosen, E.; Ullman, D.; Phillips, E.; and Tellex, S.
\newblock 2018.
\newblock {ROS reality: A virtual reality framework using consumer-grade
  hardware for ros-enabled robots}.
\newblock In {\em IEEE/RSJ International Conference on Intelligent Robots and
  Systems (IROS)},  1--9.

\bibitem[\protect\citeauthoryear{Williams \bgroup et al\mbox.\egroup
  }{2019}]{Williams2019deictic}
Williams, T.; Bussing, M.; Cabrol, S.; Boyle, E.; and Tran, N.
\newblock 2019.
\newblock {Mixed Reality Deictic Gesture for Multi-Modal Robot Communication}.
\newblock In {\em 14th ACM/IEEE International Conference on Human-Robot
  Interaction (HRI)},  191--201.

\bibitem[\protect\citeauthoryear{Williams, Szafir, and
  Chakraborti}{2019}]{williams2019reality}
Williams, T.; Szafir, D.; and Chakraborti, T.
\newblock 2019.
\newblock {The Reality-Virtuality Interaction Cube}.
\newblock {\em 1st International Workshop on Virtual, Augmented, and Mixed
  Reality for Human-Robot Interaction (VAM-HRI)}.

\bibitem[\protect\citeauthoryear{Yamauchi}{1997}]{Yamauchi1997}
Yamauchi, B.
\newblock 1997.
\newblock A frontier-based approach for autonomous exploration.
\newblock In {\em cira}, volume~97,  146.

\end{thebibliography}

\end{document}